\documentclass{article}
\usepackage{graphicx} 
\usepackage{amsmath}
\usepackage{hyperref}
\usepackage{multirow}
\usepackage{tabularx} 
\usepackage{booktabs}
\usepackage{caption}
\usepackage{tikz}
\usetikzlibrary{positioning}
\usetikzlibrary{arrows.meta}
\usepackage{subcaption}
\usepackage[style=numeric]{biblatex}
\title{The Character Error Vector: Decomposable errors for page-level OCR evaluation}
\author{Jonathan Bourne, Mwiza Simbeye, Joseph Nockels}
\date{April 2026}

\begin{document}

\maketitle

\begin{abstract}
   The Character Error Rate (CER) is a key metric for evaluating the quality of Optical Character Recognition (OCR). However, this metric assumes that text has been perfectly parsed, which is often not the case. Under page-parsing errors, CER becomes undefined, limiting its use as a metric and making evaluating page-level OCR challenging, particularly when using data that do not share a labelling schema. We introduce the Character Error Vector (CEV), a bag-of-characters evaluator for OCR. The CEV can be decomposed into parsing and OCR, and interaction error components. This decomposability allows practitioners to focus on the part of the Document Understanding pipeline that will have the greatest impact on overall text extraction quality. The CEV can be implemented using a variety of methods, of which we demonstrate SpACER (Spatially Aware Character Error Rate) and a Character distribution method using the Jensen-Shannon Distance. We validate the CEV's performance against other metrics: first, the relationship with CER; then, parse quality; and finally, as a direct measure of page-level OCR quality. The validation process shows that the CEV is a valuable bridge between parsing metrics and local metrics like CER. We analyse a dataset of archival newspapers made of degraded images with complex layouts and find that state-of-the-art end-to-end models are outperformed by more traditional pipeline approaches. Whilst the CEV requires character-level positioning for optimal triage, thresholding on easily available values can predict the main error source with an F1 of 0.91. We provide the CEV as part of a Python library to support Document understanding research.

\end{abstract}

\section{Introduction}
\label{sec:intro}

Document Understanding is a process that converts printed or digital media into a machine-readable format. Although Document Understanding is complicated and nuanced, it can be broadly broken into two sub-processes. Document Layout Analysis (DLA): What are the elements on the page, and where are they positioned? Optical Character Recognition (OCR): What does the text on the page say?
Both of these components of Document Understanding have gone through rapid development over the last few years due to the `AI Boom', with many new DLA models \cite{zhao_doclayout-yolo_2024, livathinos_advanced_2025, sun_pp-doclayout_2025}, OCR models \cite{li_trocr_2022, cui_paddleocr_2025, jaidedai_easyocr_2026, cui_pp-ocrv5_2026, paruchuri_surya_2025} and an new advanced class of end-to-end models \cite{wei_general_2024, bai_qwen25-vl_2025, poznanski_olmocr_2025, li_dotsocr_2025, ibm_granite_2025, duan_glm-ocr_2026} which combines the DLA and OCR stages into one.

While these technologies have dramatically increased the quality and speed of data extraction from images of text, two pertinent questions are: `What does ``quality" mean?' and `How to unite measures of parsing quality with OCR quality?' Within the field of Document Understanding, there are many evaluation metrics, although the F1 and CER/Edit distance are by far the most popular.

In DLA, the F1 score and Mean Average Precision (mAP) are used almost exclusively to measure how well models have located the ground-truth regions that define the page elements to be identified. However, recent research proposes that there is a triadic relationship between a dataset, a model and the evaluation metric which is applied \cite{bourne_cote_2026}. It was shown that the F1 score has a critical weakness in its ability to evaluate models in the common case where the granularity of the groundtruth does not match the granularity of the prediction. This results in a `pragmatic failure' in  which the metric misinterprets the performance of the model on the data. 

Within the field of OCR, there is more variability; however, the edit distance \cite{levenshtein_binary_1966} and the Character Error Rate (CER) are used in the vast majority of cases. These two approaches measure the number of changes required to convert the observed text string into the groundtruth string, with the CER being the edit distance normalised by the total number of characters in the ground truth. In this paper, we will refer to the CER. Whilst the CER provides practical and interpretable insight into OCR quality, it also has some well-known drawbacks which can limit its use. A key issue is that it can produce misleadingly large errors when long sub-sequences are transposed within a greater text \cite{hermansson_tracking_2016}. This typically occurs in complex multi-column layouts where the reading order is non-trivial or even ambiguous. 

As a result of these issues, more sophisticated approaches to measuring OCR quality have been developed. These approaches generally focus on taking a more flexible approach to the order of sub-sequences within the text, to give a realistic interpretation of OCR quality that is distinct from any reading order issues produced by the parsing module \cite{hermansson_tracking_2016, clausner_flexible_2020, hwang_disgo_2023, shu_computer_2024}. However, these methods still assume that the parsing of the text is still reasonable enough that sequences are still an appropriate way to measure performance; if text has been poorly parsed, the sequence becomes meaningless and the CER is undefined. It has been shown that, on complex documents, the DLA parser can break down in various ways \cite{bourne_cote_2026}, breaking the sequential logic making the CER undefined. This is distinct from the pragmatic failure of the F1, and is rather a category error of application as, in such a context, the CER itself has no meaning.

To resolve the tension between page parsing and OCR, we introduce a bag-of-characters approach that unifies parsing and OCR quality. Bag-of-characters metrics are typically inferior to sequential metrics like CER, as text is inherently sequential. However, we trade sequential analysis for spatial awareness that is robust to parsing failures. What is more, by representing the text as a character vector, we open up the possibility of treating the text as a distribution, and decomposing the error into its constituent parts. This decomposition allows for identifying whether the parsing component or the OCR component is responsible for the majority of the error. The bag-of-characters approach has the additional advantage in that it does not require a specific labelling schema such as ALTO \cite{stehno_metaeautomated_2003}, PAGE \cite{pletschacher_page_2010},  TEI \cite{tei_tei_2025}, etc, or reading order, which is challenging to extract in itself \cite{wang_layoutreader_2021,quiros_reading_2022, qiao_reading_2024, li_monkeyocr_2026}. 

Our focus on modular processes and decomposable errors harks back to the 1990s when content-holding organisations sought to recognise newly scanned material \cite{prescott_why_2018}. At that time, OCR approaches and their analysis were biologically inspired, with terms like ‘mimicking human behaviour’ and ‘human-like intelligence’ appearing frequently in the literature \cite{tanprasert_thai_1996, mani_application_1997, fan_genetic_1997}. Such work retained the understanding that human reading is rarely linear, and therefore that accurate OCR requires spatially aware approaches. When reading, humans jump between different interpretive layers of meaning \cite{terras_image_2006, tarte_interpreting_2012}. In developing a spatially aware OCR quality metric, we return to these more traditional discussions around OCR development and document understanding. Our decomposable metrics, which clearly disentangle CER and page-parsing errors, thus follow a long-standing discourse on how humans parse and interpret written text.

\subsection{Contribution}

We introduce the `Character Error Vector' (CEV), a bag-of-characters class of metrics for schema-agnostic whole-page OCR quality measures. The Character Error Vector decomposes to allow triage of the most significant source of errors. In this paper, we use two classes of discrete vector difference measures, the CER-like Count-based SpACER, and the family of measures based on distributional divergences, which we call the Character Distribution Divergence (CDD).

Concretely, our contributions are as follows.

\begin{enumerate}
    \item \textbf{SpACER}: The Spatially Aware Character Error Rate, an interpretable spatial analogue to the CER.
    \item \textbf{The Character Distribution Divergence (CDD)}: A family of metrics measuring global character extraction quality. Can be implemented using any divergence measures appropriate for characters or words.
    \item \textbf{Python Implementation}: We provide documented implementations of these approaches integrated into the \verb|cotescore| python library.
\end{enumerate}

\section{Character Error Vector}

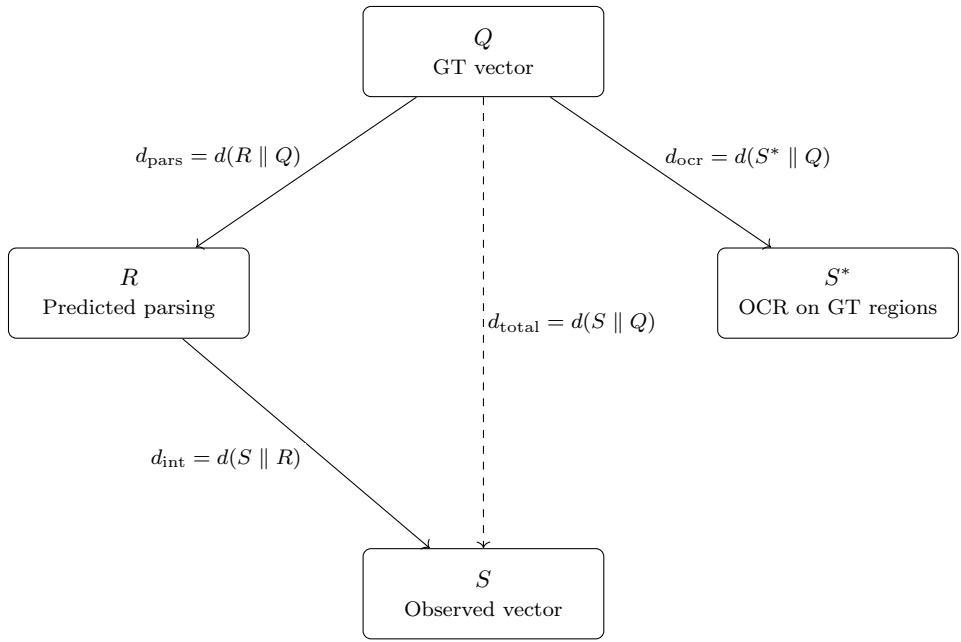
\begin{figure}[t]
\centering
\begin{tikzpicture}[
    node distance=2cm and 1.5cm,
    box/.style={
        rectangle, draw, rounded corners=3pt,
        minimum width=3.2cm, minimum height=1.2cm,
        align=center, font=\small
    },
    every edge/.style={draw, ->, >=stealth, thick},
    lbl/.style={font=\footnotesize, fill=white, inner sep=1.5pt}
]
% Nodes
\node[box] (Q) {$Q$ \\ \footnotesize GT vector};
\node[box, below left=of Q] (R) {$R$ \\ \footnotesize Predicted parsing};
\node[box, below right=of Q] (Sbar) {$S^*$ \\ \footnotesize OCR on GT regions};
\node[box, below=6cm of Q] (S) {$S$ \\ \footnotesize Observed vector};
 
% Edges
\draw[->] (Q) -- node[lbl, above left]  {$d_{\text{pars}} = d(R \parallel Q)$} (R);
\draw[->] (Q) -- node[lbl, above right] {$d_{\text{ocr}} = d(S^* \parallel Q)$} (Sbar);
\draw[->] (R) -- node[lbl, below left]  {$d_{\text{int}} = d(S \parallel R)$} (S);
\draw[->, dashed] (Q) -- node[lbl, right] {$d_{\text{total}} = d(S \parallel Q)$} (S);
 
\end{tikzpicture}
\caption{CDD decomposition when character position information is available. Solid arrows denote component differences; the dashed arrow denotes the total end-to-end divergence.}
\label{fig:cdd_decomposition}
\end{figure}

In a document OCR pipeline, there are two key stages. The parsing and OCR stages. Both elements are sources of error that need to be measured. In this process, we define three types of error. Parsing, OCR, and interaction. These three error types are described in table \ref{tab:cdd_decomposition}. The Table Shows how the error types relate to the underlying vectors. The vectors themselves are shown in Table \ref{tab:distribution_types}, whilst a schematic of the relationships is shown in Figure \ref{fig:cdd_decomposition}. 

When character positions are known, or can reasonably be inferred, the three error types can be directly calculated and the most important contributor to the total error identified. This allows for mitigating actions, such as additional training data or replacing either the Parsing or OCR component of the pipeline, to reduce the total error $d_{total}$. It should be noted that as the values are produced from high-dimensional vectors the sub-components are not additive and so $d_\textrm{total} \neq d_\textrm{ocr} + d_\textrm{pars} + d_\textrm{int}$, for all practical cases.

\begin{table}[t]
\centering
\caption{CDD decomposition into parsing and OCR error components when character position information is available.}
\label{tab:cdd_decomposition}
\begin{tabular}{@{}lllp{6.2cm}@{}}
\toprule
Component & Difference & Comparison & Description \\
\midrule
Parsing & $d_{\text{pars}}$ & $d(R \parallel Q)$ &
 Errors caused by page regions being missed or captured more than once due to overlapping detections. \\
OCR & $d_{\text{ocr}}$ & $d(S^{*} \parallel Q)$ &
  Transcription errors resulting in deviation from the true character distribution, isolated on GT regions. \\
Interaction & $d_{\text{int}}$ & $d(S \parallel R)$ &
  OCR errors which are model dependent but on the parsing of the document image. \\
\addlinespace
Total & $d_{\text{total}}$ & $d(S \parallel Q)$ &
  Combined character error encompassing both parsing and OCR errors. \\
\bottomrule
\end{tabular}
\end{table}

\begin{table}[t]
\centering
\caption{Character vectors used in the CDD decomposition.}
\begin{tabular}{@{}clp{3.5cm}@{}}
\toprule
Symbol & Definition & Error captured \\
\midrule
$Q$ & GT character vector &  No Error \\
$R$ & Vector from the predicted parsing on GT characters & Parsing error only \\
$S^*$ & Vector created by performing OCR on GT regions & OCR error only \\
$S$ & Vector created by performing OCR on predicted regions & Combined OCR and Parsing \\
\bottomrule
\end{tabular}
\label{tab:distribution_types}
\end{table}

Functionally, we see that by sacrificing the CER's sequential awareness and replacing it with spatial awareness, we can move beyond the constraints of the ground-truth bounding box to a more generalised and flexible measure of OCR quality. 

\subsection{Assumptions}
\label{sec:assumptions}
The Character Error Vector makes several assumptions

\begin{itemize}
    \item Large number of characters: The CEV is designed to be used at page level, which typically contains hundreds or thousands of characters, at this scale the distributional approximation is robust.
    \item $S$ is not a permutation of $R$: This can occur when $R$ is the stationary distribution of the transmission matrix $T_{ocr}$. It would result in all the right characters in all the wrong places, but be undetectable as the distribution would be correct. This is unlikely due to the small number of stationary distributions for a given transmission matrix (if any), and the fact that the distributions vary by language, whilst the OCR errors are primarily visual, therefore script dependent.
    \item $S$ is not a permutation of $Q$: Same as above, but even less likely due to the non-linear geometric transform that takes place in $T_{parse}$.
\end{itemize}

With these assumptions, the CEV would be proportional to CER under perfect parsing.

\subsection{Choice of underlying metric}

Although the CEV can be applied with any discrete vectors, in this manuscript, we provide two implementations with distinct properties, The SpACER and the Jensen-Shannon Distance. These two approaches are introduced in the following sub-sections. For practitioners, SpACER is most familiar as analogous to the CER, providing an easily interpretable measure of the magnitude of the error. The Jensen-Shannon Distance provides a better measure of the change in the shape of the distribution. 

\subsubsection{SpACER}

Below, we define the Spatially Aware Character, Error Rate (SpACER). Similar to the WER, the Spatially Aware Error Rate (SpAWER) is identical to the SpACER except with words instead of characters. For the purposes of this definition, we will refer only to the SpACER.

Firstly, we find the error between the groundtruth and predicted character counts using the L1 norm

\begin{equation}
    \hat{E} = \left\| g - p \right\|_1
\end{equation}

Where $g$ ground truth character count vector, and $p$ is the predicted character count vector. $\hat{E}$ is therefore equivalent to the lower bound of the error for insertions and substitutions in the CER calculation. It is the lower bound, as pure character swaps are unobservable and only the total difference per character can be measured. In the case where the third CEV assumption is violated and the errors are pure swaps $g=p$ and the apparent error is 0.

We then calculate the SpACER as follows

\begin{equation}
    \textrm{SpACER} = \frac{D + \hat{E}}{2C}
\end{equation}

Where $C$ is the character count in the ground truth $C = \left| g \right|$, and D is the total number of deletions relative to the ground truth character count, such that

\begin{equation}
    D = \textrm{max}\left( 0,   \left| g \right| - \left| p \right|
\right)
\end{equation}

As can be seen, this approach blends the normalisation of the Total Variation Distance with the Deletion sensitivity of the CER. This creates a value that is symmetric relative to deletions and insertions and, like the CER, can extend beyond 1 if the errors are high enough. 

The SpACER can be calculated at both bounding box and page level, providing micro and macro values. At the micro level, the Deletions counts are summed across all image predictions, whilst at the macro level, the total Deletions at the page level is all that is used; this means that the micro provides a clearer understanding of OCR quality, as deletions and insertions between prediction boxes do not cancel out. However, this requires more granular ground truth data and is not usable for end-to-end models analysing an entire page. 

The definition of the micro SpACER is

\begin{equation}
    \textrm{SpACER}_{micro} = \frac{\sum^n_j D_j + \hat{E}}{2C}
\end{equation}

Where $j$ is the jth prediction produced by the page parsing model and $D$ is the sum of all $D_j$ across all $n$ predictions. The macro SpACER simply uses all predicted and groundtruth character at page level

SpACER and CER become identical when two conditions are met: The parsed region must include only the characters of the ground truth; the transcription errors must have no swap substitutions. As an example, if an ``a" is substituted for a ``b", then ``b" cannot substitute for an ``a".  Similar to CER, SpACER is not a true metric as reversing the groundtruth and observed value does not produce the same result, due to the $C$ in the denominator. In contrast, dropping the $C$ term would produce a true distance metric, the Spatially Aware Character Distance, analogous to the relationship between CER and Edit Distance.

\subsubsection{The Jensen-Shannon Distance}

The Jensen-Shannon Distance (JSD) represents the distribution-based element of the CEV known as the Character Distribution Divergence (CDD). The CDD can be any discrete divergence or distributional distance. We choose the JSD because, unlike SpACER, it is a proper distance metric, meaning it is symmetrical and satisfies the triangle inequality; in addition, it is bounded between 0 and 1. This contrasts nicely with the properties of SpACER.
The JSD is the square root of the Jensen-Shannon divergence, which itself is derived from the Kullback-Leibler divergence. Both of these measures of distributional difference are widely used across the fields of NLP and Machine Learning \cite{pillutla_mauve_nodate, lu_diverging_2020, mehri_word_2015, lafferty_document_2017}. As such, we can derive this CDD implementation using Shannon entropy in bits.

\begin{equation}
    H(X) = -\sum \textrm{P}(x)log_2 \textrm{P}(x)
\end{equation}

\begin{equation}
    \textrm{CDD}_\textrm{JSD}(S||Q)^2 = \textrm{JSD}(S||Q)^2 = \textrm{H}(M) - \frac{1}{2}(\textrm{H}(S)+\textrm{H}(Q))
\end{equation}

\begin{equation}
    \textrm{CDD}_\textrm{JSD}(S||Q) = \textrm{JSD}(S||Q) 
    \end{equation}

\begin{equation}
    M = \frac{1}{2}(S + Q)
\end{equation}

We would expect the JSD to return larger differences when rare characters are inserted or deleted due to OCR. This is valuable, as OCR systems, particularly older non-LLM systems, often introduce symbols due to page noise, which can make reading challenging.

\section{Method}

\subsection{Data}
\label{sect:data}

The dataset comes from the `The Spiritualist' a 19th-century newspaper focused on the supernatural \cite{nls_spiritualist_2019}. In the original dataset the OCR was low quality, we high-quality transcription sub-set obtained using Transkribus \cite{kahle_transkribus_2017} with manual correction \cite{nockels_spiritualist_2026}. We use this dataset as archival newspapers are known for their difficulty in document understanding, due to their complex layouts and degraded images \cite{cordell_q_2017}.

Whilst the original contained 50 pages we use 49 as page 17 of the dataset is so degraded as to be unreadable, this links to general ideas of data loss from ephemeral collections \cite{kilbride_saving_2015}. A dataset of this size is sufficient to validate the CEV and demonstrate the kind of behaviours that may be observed. 

The sub-set used the ALTO XML format, and did not distinguish between text types (e.g., headers and body text) or provide column information or reading order. To enable more robust analysis of the metrics, we enriched the dataset with text class labels, column information, and reading order. We also grouped the text into simple Structural Semantic Units \cite{bourne_cote_2026}. Dataset properties are shown in Table \ref{tab:dataset_info}, more detail is provided on the enrichment process in the supplementary material Section 2.

\begin{table}[h!]
\centering
\caption{Details on The Spiritualist dataset used in this paper}
\label{tab:dataset_info}
\begin{tabular}{|l|c|}
\hline
\textbf{Property} & \textbf{Value} \\
\hline
Images      & 49 \\
\hline
Lines       & 19,691 \\
\hline
Words       & 175,548 \\
\hline
Granularity & Word level \\
\hline
Region Type & Polygon \\
\hline
\end{tabular}
\end{table}

\subsection{Models}
\label{sec:meth_model_comp}

To explore SpACER and the CDD for OCR quality analysis, we create Document text extraction pipelines by combining Page Parsing and OCR models. 

For the parsing models we will be using Doclayout-yolo \cite{zhao_doclayout-yolo_2024}, Heron \cite{livathinos_advanced_2025}, and PP-Doclayout \cite{sun_pp-doclayout_2025}.  Three sizes of PP-Doclayout are used to test the impact of parameter count. Details of the models can be seen in Table \ref{tab:layout_models}. 

For the OCR models we will be using, the following models. Tesseract \cite{smith_overview_2007} is a classical OCR model that is still in common use and performs well on degraded images. EasyOCR \cite{jaidedai_easyocr_2026} is a popular convolutional neural network; it is paired with the Heron parser in the Docling framework. TrOCR \cite{li_trocr_2022} is an early transformer-based model. PaddleOCR \cite{cui_paddleocr_2025}, a cutting-edge model that is typically part of the Paddle Paddle Document Understanding pipeline, which includes the PP-Doclayout models.

It should be noted that OCR models typically only handle a single line of text; as such, the text needs to be split up at line-level, EasyOCR, Tesseract, and PaddleOCR all handle this internally; however, TrOCR does not. We run TrOCR using Tesseracts internal line splitter. Whilst the quality of line-splitting introduces a new source of uncertainty, this is a necessity for all systems, and further exploration of this area is beyond the scope of the paper.

\begin{table}[htbp]
\centering
\caption{Comparison of OCR Models}
\label{tab:ocr_models}
\begin{tabular}{lccc}
\toprule
\textbf{Model} & \textbf{Type} & \textbf{Parameters} & \textbf{Framework} \\
\midrule
PaddleOCR & CNN    & 23.9M   & PaddlePaddle \\
TrOCR     & Transformer  & 334M   & PyTorch      \\
EasyOCR   & CNN    & 13.4M  & PyTorch      \\
Tesseract & Traditional  & —      & C++          \\
\bottomrule
\end{tabular}
\end{table}

\begin{table}[htbp]
\centering
\caption{Comparison of Document Layout Analysis Models}
\label{tab:layout_models}
\begin{tabular}{lccc}
\toprule
\textbf{Model} & \textbf{Type} & \textbf{Parameters} & \textbf{Framework} \\
\midrule
DocLayout-YOLO & CNN & 15.4M & PyTorch \\
Heron & Transformer & 42.9M & PyTorch \\
PP-DocLayout-L & Transformer  & 30.94M & PaddlePaddle \\
PP-DocLayout-M & Transformer & 5.65M & PaddlePaddle \\
PP-DocLayout-S & Transformer & 1.21M & PaddlePaddle \\
\bottomrule
\end{tabular}
\end{table}

Beyond the classical, composable pipeline of document parsers followed by OCR extraction, many modern OCR systems are end-to-end. These frameworks are typically based entirely on vision-language models (VLMs), with no distinction between the parsing and OCR stages. These models have rapidly become state-of-the-art, but often at the cost of interpretability. As a comparison, we use three such models, olmOCR \cite{poznanski_olmocr_2025}, dots.MOCR \cite{li_dotsocr_2025}, and Granite Docling \cite{ibm_granite_2025}, which is also part of the Docling framework.

\begin{table}[htbp]
\centering
\caption{Comparison of End-to-End OCR Models}
\label{tab:e2e_ocr_models}
\begin{tabular}{lccc}
\toprule
\textbf{Model} & \textbf{Type} & \textbf{Parameters} & \textbf{Framework} \\
\midrule
olmOCR       & Transformer  & 7B    & PyTorch \\
dots.MOCR     & Transformer  & 7B    & PyTorch \\
Granite-Docling  & Transformer  & 258M  & PyTorch \\
\bottomrule
\end{tabular}
\end{table}

All models will be run using default parameters; none of the models were trained on the evaluation data described in section \ref{sect:data}.

\subsection{Evaluation Metrics}

Beyond the CDD metrics described previously, we will also use the CER, COTe score, and F1. These three metrics will be used to analyse both the parsing and direct OCR quality of the models, and linking this to the expressiveness of the CEV metrics.  Across all OCR evaluation the text
will be normalised by setting all text to lower case, standardising punctuation symbols, removing double spacces, etc.

The CER is an obvious choice due to its close alignment with the SpACER. The CER identifies character sequence errors, classifying them as either Substitutions, Deletions, or Insertions, the sum of which is divided by the total number of characters in the ground truth.

\begin{equation}
    \frac{\textrm{Substitutions}+\textrm{Deletions}+\textrm{Insertions}}{\textrm{Total characters}}
\end{equation}

The Coverage, Overlap, Trespass, Excess score, \cite{bourne_cote_2026}, provides a decomposable framework identifying page-parsing errors which affect the semantic content of the text extracted. Coverage ($\mathcal{C}$) defines how much of the ground truth regions on the page are covered, Overlap ($\mathcal{O}$) identifies areas which are covered by more than one prediction, and Trespass  ($\mathcal{T}$) describes predictions which cross from one semantic ground truth region to another, Excess a support metric measures how much of the page not covered by a ground truth region was also captured. The cote is calculated using

\begin{equation}
    \textrm{COTe score} = \mathcal{C} - \mathcal{O} - \mathcal{T} 
\end{equation}
The COTe score is maximised at 1 where Coverage is 100\% whilst Overlap and Trespass are both 0.

Despite it's drawbacks \cite{bourne_cote_2026, heo_led_2025}, the F1 is by far the most popular way to measure page-parsing quality and so is included for completeness.

\begin{equation}
    F_1 = 2 \cdot \frac{\text{Precision} \cdot \text{Recall}}{\text{Precision} + \text{Recall}}
\end{equation}

Where Precision and Recall are as shown below
\begin{equation}
    \text{Precision} = \frac{\text{TP}}{\text{TP} + \text{FP}} = P,
\quad
\text{Recall} = \frac{\text{TP}}{\text{TP} + \text{FN}} =R
\end{equation}
 where TP is True Positive, FP is False Positive, and FN is False Negative.

\subsection{Measuring positional uncertainty}
 
To perform a full decomposition, the CEV methods presented in this paper rely on the positions of characters in the image. However, this is rarely available in ground-truth datasets \cite{clausner_flexible_2020}. Normally, datasets are available at best at word-level or line-level, since most traditional OCR metrics operate at line-level.  If we infer the position of the characters using equal spacing per character (mono-spacing), how does this degrade the reliability of the CEV? 

To understand this degradation, we can take a random crop of text and compare the characters within the crop when using groundtruth vs inferred positions. By identifying each character with its unique xy coordinate, we can definitively say which characters are not in the crop, which should be, or are in the crop when they should not. This allows us to measure the character error using the CER. Using this approach, we would expect to see the CER increase proportionally with the average number of letters per bounding box.

To conduct this test, we generate a large amount of synthetic text using the approach from \cite{bourne_scrambled_2025}, and generate 10 pages for each combination of the Font, Layout and Columns shown in table \ref{tab:synth_combos}, page size is held constant at 280mm x 430mm. This produces 450 pages of text in total. The fonts were chosen for their high visual contrast; the three alignments present different spacing issues, whilst the columns represent the different types of layout in which text is commonly found in printed media. For some examples of this data please see Supplementary Material Section 3

\begin{table}[h!]
\centering
\caption{Parameters used in text generation process}
\label{tab:synth_combos}
\begin{tabular}{p{3.5cm}p{2cm}c}
\toprule
\textbf{Font} & \textbf{Layout} & \textbf{Columns} \\
\midrule
 Ewert, Great Vibes, Kablammo, Roboto, UnifrakturMaguntia  & Centered, Justified, Left aligned & 1, 2, 3 \\
\bottomrule
\end{tabular}
\end{table}

To obtain a robust estimate of CER degradation, we randomly sample crops from the page. These crops are sized for all 25 combinations of 10\%, 20\%, 30\%, 40\%, 50\% of image height and width. Each combination is repeated 30 times. Across all images, this creates a matrix of 337,500 total samples, with 13,500 samples per size combination. Using a matrix, we are able to check the CER degradation across a wide range of scenarios, at word, line and paragraph level \footnote{Technically this is the Structural Semantic Unit \cite{bourne_cote_2026}, but for the given setup is very similar to paragraphs}.

It should be noted that real DLA models, even when performing poorly, respond to the page structure, unlike the random crops used here. This means our crops should be considered worst-case when the predicted regions are entirely unaligned with the document itself.

\subsection{Empirical Validation}

As the CEV is a new approach to measuring OCR quality, it must be clearly validated. We do this at three different levels. 

The first level is comparing the CEV directly as a CER equivalent using the perfectly parsed text blocks of the ground truth regions. This perfect parsing makes the spatial element of the CEV irrelevant, and the CER becomes the ideal metric. As a result, the CEV should return a value that is proportional to the CER, where poorer OCR quality results in higher CER and SpACER scores. 

The second test compares CEV against parsing quality. We need to know whether the CEV accurately reflects how well the page has been parsed. We do this by measuring the predicted parse quality of the parsing models, using F1 or COTe score, and comparing with the page-level CEV score under perfect OCR. We should expect to see that as parsing quality reduces, the CEV score increases. 

Finally, having validated the CEV controlling for either the spatial or OCR element, we test controlling for neither. This is the most challenging element due to the complex structure of newspapers. We handle the issue in this case by exploiting the pre-existing labelled ground truth produced by Transkribus, which includes text box ID and columns. We use this information to construct an approximate reading order, then concatenate the ground truth texts for each page into a single ordered text string. We then match the parsed boxes to the ground truth boxes by position to create a parallel reading order, and then concatenate the OCR results into a single string. 
The approximate ordering allows us to perform a CER analysis, which, given the volume of page model combinations, should indicate whether there is a correlation in CEV at the page level. 

\subsection{Pipeline Comparison}

Using the model pipelines, we will combine the parsing and OCR models together and decompose the errors to better understand how the CEV works and its interpretation. We are interested in understanding how robust the approach is to different kinds of error within the pipeline and the general pragmatic competence of the system \cite{bourne_cote_2026} in correctly interpreting the pipeline performance.

\subsection{End-to-end models}

As end-to-end models are increasingly popular, within the field of text extraction, we will apply the three end-to-end models mentioned in Section \ref{sec:meth_model_comp}, comparing their performance to the pipeline models.

\subsection{Low Information Analysis}

One of the major advantages of the CEV is the ability to decompose errors, identifying whether the biggest performance boost will come from improving either Parsing or OCR elements. However, a full decomposition is not always possible; for example, the data may not have character positions, or the architecture of the model may not provide parsing information, which is particularly relevant when using end-to-end models.

As such, we will test the CEV's ability to triage between OCR and parsing as the main sources of error, that is if $d_{pars} \leq d_{ocr}$. We will use simple threshold of the ratio of $d_{ocr}$ over $d_{total}$

\section{Results}

The results are broken into three sections. First, we measure the impact of inferring character position at different granularities, then we validate the CEV using three comparison tests. We then compare the different document pipelines and the error decompositions they produce. Following this we provide a qualitative analysis of OCR as a result of parsing error. Next we analyse end-to-end models. Finally, we explore whether the dominant error source can be predicted without character position.

\subsection{Spatial granularity}

\begin{figure*}
    \centering
    \includegraphics[width=\linewidth]{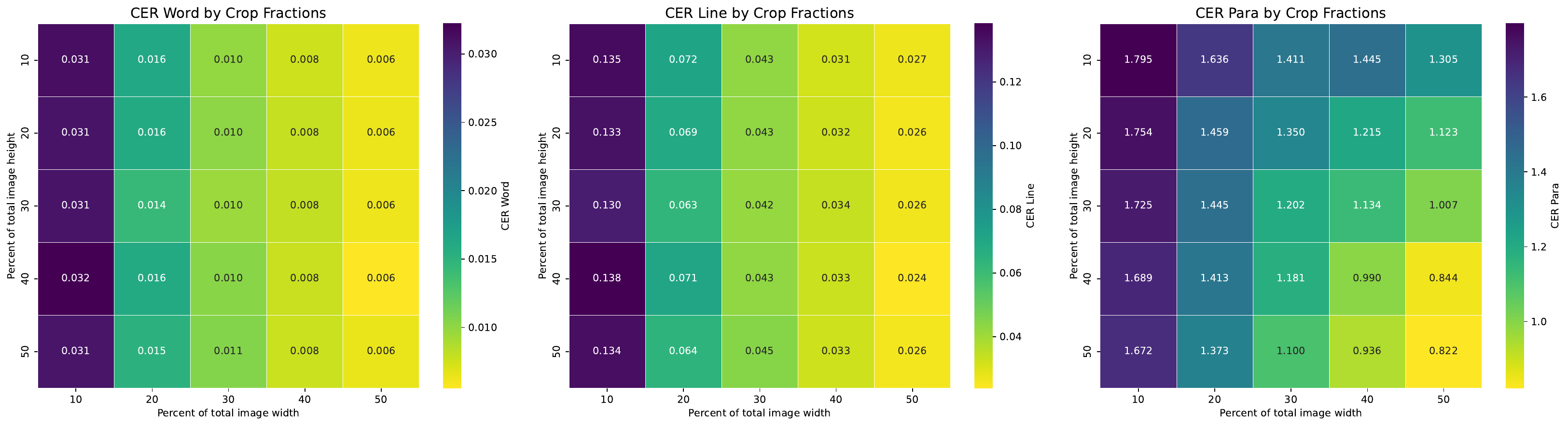}
    \caption{Analysis of the impact of spatial uncertainty on the character position of the ground truth. We see that increasing the prediction width increases performance, but that paragraph-level inference is unusable at any scale.}
    \label{fig:spatial_granularity_error}
\end{figure*}

We analyse the effect of reduced uncertainty on character position on the page. Figure \ref{fig:spatial_granularity_error} shows the results of 377 thousand samples across 450 pages of text. It is clear that there are significant differences between inferring at the word (1.4\% CER), line (6\% CER), and paragraph levels (132\% CER), with the mean and median being generally similar. Changing the box height affects only the paragraph-level inference, since that is the only one that spans multiple lines. Very small boxes have relatively large CER because they have fewer characters in total; they are therefore more sensitive to changes. The error roughly halves for word, and line-level when box width increases from 10\% to 20\%, as the width doubles. 

Text alignment did not affect the CER. This is because at the word level there are no spaces to be aligned, and at the line level the total amount of space doesn't change it is just redistributed. Similarly, the number of columns had no effect on the CER; this was because the predictions were randomly placed and unrelated to the text layout. Typically, prediction boxes are as wide as, or wider than, the column width. This means that, with real predictions, the error due to spatial uncertainty will be inversely proportional to the number of columns. 

The Spiritualist typically has 3 columns, which, including margin, would make the prediction regions roughly 25\% of the page width, so the expected CER from uncertainty would be 1.3\% and 5.5\% for word and line inference, respectively, paragraph inference being unusable. As such, using the word-level inference is a reasonable choice for the following experiments.

\subsection{Empirical Validation of Spacer and CDD}

Having established that word-level inference is acceptable, we now validate SpACER and CDD at three levels: the ability to measure OCR errors, the ability to measure parsing errors, and finally the ability to measure whole page text extraction quality.

\subsubsection{Validating ability to measure pure OCR errors}

First, we validate that, for a given bounding box, SpACER and CDD reflect the sequential CER score. We do this using the Spearman correlation, as CER-like metrics often have non-normal distributions. The results are shown in Table \ref{tab:cer_spearman}. EasyOCR has the lowest correlation with a moderate value of 0.66; in contrast, PaddleOCR, Tesseract, and TrOCR all have very strong correlations of 0.95, 0.91, and 0.96, respectively. The CDD has lower scores because it weights rare characters more heavily than CER and SpACER. However, apart from TrOCR, which has by far the worst OCR quality, the CDD still shows moderate to strong correlation. These findings confirm that at its most granular, SpACER and CDD do represent transcription errors. 

\begin{table}[t]
\caption{Spearman correlation ($\rho$) between CER and $d_\text{ocr}$ for SpACER and CDD, computed over GT regions.}
\label{tab:cer_spearman}
\begin{tabular}{lcc}
\toprule
 & SpACER $\rho$ & CDD $\rho$ \\
\midrule
Easyocr & 0.657 & 0.542 \\
PaddleOCR & 0.945 & 0.680 \\
Tesseract & 0.910 & \textbf{0.704} \\
TrOCR & \textbf{0.963} & 0.267 \\
\bottomrule
\end{tabular}
\end{table}

\subsubsection{Validating ability to measure parsing errors}
We then look at the correlation between SpACER and the COTe score. We use the COTe score due to the risk of CER being undefined, and, as shown in Table \ref{tab:d_pars}, mAP cannot be used because it completely failed to measure model performance, aligning with the findings of \cite{bourne_cote_2026}. Table \ref{tab:dpars_cote_spearman}, shows that parsing quality is inversely proportional to SpACER and CDD scores. This is to be expected as improved parsing means more accurate character capture. Heron, PPDoc-L, and YOLO all have moderate to strong negative correlation, whilst PPDoc-m and PPDoc-s have weak negative/positive correlation. The significantly different scores of PPDoc-M and PPDoc-S are due to their poor parsing quality. The parsing issue arises because these models make degenerate predictions that cover almost the entire page; this is examined further in Section \ref{sec:pipeline}. Interestingly, the CDD is more extreme on every measure than SpACER; this may be due to the distribution's shape being particularly sensitive to geometric errors associated with parsing. However, SpACER and CDD have the same ranking as no new symbols are introduced, and Overlap is low. 

\begin{table}[t]
\caption{Spearman correlation ($\rho$) between per-page $d_\text{pars}$ and COTe score by parsing model. Negative $\rho$ indicates that higher COTe (better parsing geometry) corresponds to lower parsing error, as expected.}
\label{tab:dpars_cote_spearman}
\begin{tabular}{lcc}
\toprule
 & SpACER $\rho$ & CDD $\rho$ \\
\midrule
Heron & -0.577 & -0.742 \\
PPDoc-L & \textbf{-0.742} & \textbf{-0.809} \\
PPDoc-M & -0.275 & 0.285 \\
PPDoc-S & 0.181 & 0.250 \\
YOLO & -0.522 & -0.649 \\
\bottomrule
\end{tabular}
\end{table}

\subsubsection{Validating ability to measure whole page extraction quality}
\begin{figure}
    \centering
    \includegraphics[width = \linewidth]{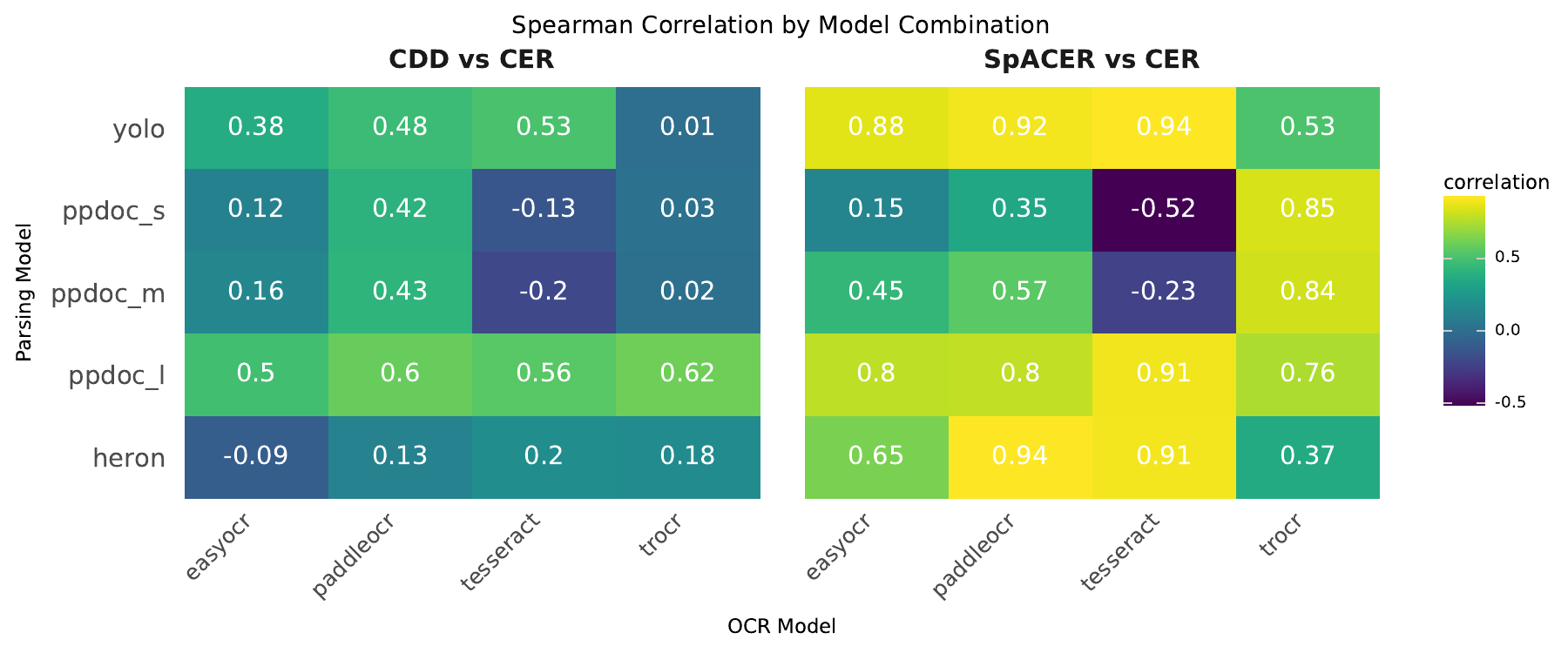}
    \caption{We can see that SpACER has good correlation with CER at page level, indicating that they are both measuring the same thing. The CDD has low correlation, as it measures change in shape of the distribution not the error rate.}
    \label{fig:CEV_CER_correlation}
\end{figure}

We now compare SpACER, the CDD, and the CER at page level by reconstructing the approximate text order and calculating CER. From Figure \ref{fig:CEV_CER_correlation} we see that, with caveats for PPDoc-m and PPDoc-2,  SpACER and CER are generally moderately to strongly correlated. This indicates that SpACER is, in fact, a good spatial proxy for CER. In contrast, the CDD has a low to moderate relationship with CER. This is intuitive, as the CDD measures the change in the distribution's shape, not the error rate. However, as the CDD and SpACER share the same underlying mechanism of the CEV, and we can see that SpACER is measuring character error rate, we can infer that the CDD reflects genuine changes in the shape of the character distribution at page level. We also see that CER correlation drops significantly for PPDoc-m and PPDoc-s due to their degenerate parsing (See section \ref{sec:pipeline}).

By validating the performance of the CEV metrics, controlling for parsing and OCR, and finally full-page-level data extraction, we demonstrate that this suite of metrics can represent OCR quality in a meaningful way. As such, it provides a single value that represents OCR quality, integrating both global parsing and local text-extraction quality.

We now proceed to analysing the decomposition of the models in more detail.

\subsection{Pipeline Comparison}
\label{sec:pipeline}

Table \ref{tab:d_total} shows the overall results for the total OCR quality using the SpACER macro score. We find that SpACER micro and macro have similar results, and so we will only refer to SpACER macro throughout the results. 

We observe substantial interaction between the parsing and OCR models regarding the quality of results. In relation to OCR models, PaddleOCR and Tesseract perform significantly better than EasyOCR and TrOCR. When looking at the parsing models, we see that the Heron model is a strong performer, though it is outperformed by both PPDoc-M and PPDoc-S when using the PaddleOCR model. However, it is peculiar, as PPDoc-L is substantially outperformed by the smaller PPDoC models on On EasyOCR and PaddleOCR. What is more, PPDoc-M and PPDoc-S perform very badly on Tesseract and TrOCR. These irregularities can be further investigated and explained by decomposing the SpACER score and combining it with complementary metrics.

\begin{table}[t]
\caption{Page-level SpACER error ($d_\text{total}$) by parsing model (rows) and OCR model (columns). \textbf{Bold}: column best; \textbf{bold}$^*$: overall best.}
\label{tab:d_total}
\begin{tabular}{lcccc}
\toprule
 & Easyocr & PaddleOCR & Tesseract & TrOCR \\

\midrule
Heron & 0.031 & 0.009 & \textbf{0.009} & \textbf{0.036} \\
PPDoc-L & 0.049 & 0.024 & 0.029 & 0.053 \\
PPDoc-M & 0.033 & \textbf{0.006}$^{*}$ & 0.126 & 0.160 \\
PPDoc-S & 0.033 & 0.006 & 0.114 & 0.159 \\
YOLO & \textbf{0.028} & 0.015 & 0.016 & 0.037 \\
\bottomrule
\end{tabular}
\end{table}

\begin{figure*}
    \centering
    \includegraphics[width=\linewidth]{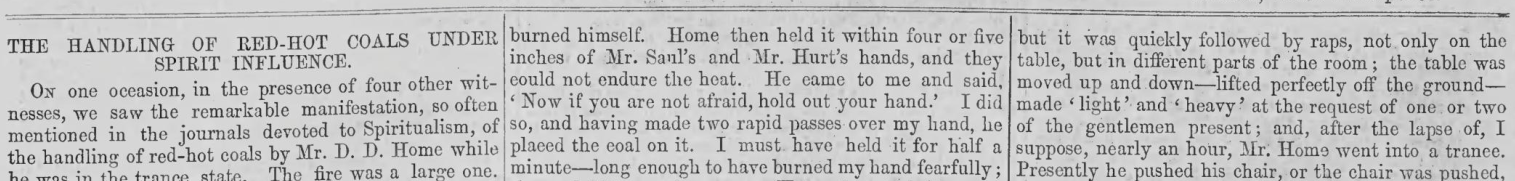}
    \caption{PPDoc-m and PPDoc-S, fail to correctly parse the images resulting in multiple columns in each crop, such as this example from image 1. All the OCR models fail to produce sensible text, but the failure mode depends on model architecture.}
    \label{fig:burnt_hands}
\end{figure*}

\subsection{Decomposition analysis}

When considering the parsing element, Table \ref{tab:d_pars} shows a similar story to the overall result shown in Table \ref{tab:d_total}. The Heron and YOLO models perform well, the PPDoc-L model performs worst, but the PPDoc-M and PPDoc-S models perform exceptionally well. However, the COTe score reveals that the smaller PPDoc models have performed very badly. Visual inspection of the results shows that PPDoc-M and PPDoc-S have produced degenerate parsings producing only a few large regions that covered the entire page. These visual finding are supported by the COTe's Trespass score which is very high for PPDoc-M (0.74) and PPDoc-S (0.74), Compared to Heron (0.0), PPDoc-L (0.03), and YOLO (0.0). This disparity between SpACER and COTe, highlights a weakness of SpACER that a single prediction which simply covers the entire image will have a perfect $d_{pars}$. 

The impact of this Trespass is discussed in Section \ref{sect:trespass_impact_analsis}.  Detailed information on the COTe scores can be found in Supplementary Material Section 4.

\begin{table}[t]
\caption{Parsing error ($d_\text{pars}$) by parsing model with COTe and mAP@0.5. SpACER macro is the primary metric; lower is better for SpACER/CDD, higher for COTe/mAP.}
\label{tab:d_pars}
\begin{tabular}{lcccc}
\toprule
 & SpACER & CDD & COTe & mAP@0.5 \\

\midrule
Heron & 0.013 & 0.008 & 0.843 & 0.074 \\
PPDoc-L & 0.025 & 0.011 & 0.749 & 0.084 \\
PPDoc-M & 0.001 & 0.003 & 0.126 & 0.000 \\
PPDoc-S & \textbf{0.001} & \textbf{0.000} & 0.264 & 0.000 \\
YOLO & 0.015 & 0.009 & \textbf{0.862} & \textbf{0.099} \\
\bottomrule
\end{tabular}
\end{table}

When considering pure OCR, Table \ref{tab:d_ocr} shows that PaddleOCR was the most robust model, getting the lowest error on SpACER, CDD, and CER. Tesseract was the second-best model in terms of SpACER and CER, but third for CDD. EasyOCR and Tesseract have higher CDD scores because they tend to insert random symbols based on visual noise, which also affects the CER score. Across the models, we see evidence that SpACER is the lower bound on the CER, with a value roughly 50\% of the measured CER at the bounding-box level.

\begin{table}[t]
\caption{OCR error ($d_\text{ocr}$) by OCR model, averaged over pages using GT regions. SpACER macro is the primary metric; lower is better.}
\label{tab:d_ocr}
\begin{tabular}{lccc}
\toprule
 & SpACER & CDD & CER \\
\midrule
Easyocr & 0.047 & 0.087 & 0.121 \\
PaddleOCR & \textbf{0.020} & \textbf{0.020} & \textbf{0.028} \\
Tesseract & 0.024 & 0.056 & 0.050 \\
TrOCR & 0.053 & 0.031 & 0.092 \\
\bottomrule
\end{tabular}
\end{table}

Table \ref{tab:d_int} shows that the error due to the interaction between parsing and OCR is primarily architecture-based, with OCR models producing very consistent errors across different parsing models (except for the Trespassing models). PaddleOCR and Tesseract produced almost 3 times fewer errors than EasyOCR and TrOCR. This is an interesting finding in itself, given the fundamentally different OCR mechanisms of PaddleOCR and Tesseract, but further investigation is beyond the scope of this paper.

\begin{table}[t]
\caption{Interaction error ($d_\text{int}$, SpACER) by parsing model (rows) and OCR model (columns). \textbf{Bold}: column best; \textbf{bold}$^*$: overall best.}
\label{tab:d_int}
\begin{tabular}{lcccc}
\toprule
 & Easyocr & PaddleOCR & Tesseract & TrOCR \\
\midrule
Heron & 0.032 & 0.012 & 0.012 & 0.035 \\
PPDoc-L & 0.029 & 0.010 & \textbf{0.012} & 0.038 \\
PPDoc-M & 0.032 & 0.006 & 0.103 & 0.158 \\
PPDoc-S & 0.033 & \textbf{0.006}$^{*}$ & 0.115 & 0.158 \\
YOLO & \textbf{0.027} & 0.014 & 0.013 & \textbf{0.028} \\
\bottomrule
\end{tabular}
\end{table}

The decomposition analysis of the SpACER score shows that, in fact, the pipelines that at first glance appear best should be disqualified. This means that PaddleOCR and Tesseract are joint first when combined with the Heron parsing model, having both got a SpACER score of 0.009.

\subsection{Analysing the impact of parsing errors on OCR}
\label{sect:trespass_impact_analsis}

We take the PPDoc-s output for page 1 of the dataset. In this case, the PPDoc-S parsing box covers multiple columns, part of which is shown in Figure \ref{fig:burnt_hands}. This parsing failure is notable as it erroneously produces low SpACER and CDD scores whilst outputting junk.
We manually compare the output OCR, shown in Table \ref{tab:burnt_hands} \footnote{TrOCR didn't output anything meaningful enough to be included}, with the ground truth text. Although ultimately all models failed to handle the cropped image correctly, how they failed reveals interesting architectural differences and implications for CEV. Tesseract produced the least accurate results; it showed high noise, failing to get even a few correctly spelt words. This is likely due to its simple model, which is highly dependent on letter recognition. EasyOCR had a much lower SpACER score. It converts small groups of words and uses a deep learning architecture, resulting in many fewer spelling errors; however, some clearly erroneous additions, such as ``\$", are still present, which cause EasyOCR to have a higher JSD. Finally, PaddleOCR, the most advanced model, produces very high-quality text, as shown by its lowest score in Table \ref{tab:d_total}; however, it produces all the right words in the wrong order, creating an unusable result. The PaddleOCR result is functionally breaking the assumption discussed in Section \ref{sec:assumptions}, not at the character level but at the word level.

\begin{table}
    \centering
        \caption{Comparison of OCR model outputs from image 1 of the dataset when using PPDoc-S parsing regions.}
    \begin{tabularx}{\textwidth}{lcX}
        \toprule
        \textbf{Model} & SpACER & \textbf{Output Text} \\
        \midrule
        EasyOCR & 0.029&thc SPIRIT INFLUENCE. inches of Mr. Saul'\$ and Mr. Hurt's hands, and table, but in difierent parts of the room the table was ON one occasiol; \\[6pt]
        PaddleOCR & 0.007 & THE HANDLING OF inches of Mr. Saul's and Mr. Hurt's hands, and they SPIRIT INFLUENCE. table, but in different parts of the room; the table was \\[6pt]
        Tesseract & 0.138 &SPIRIT INFLUEN oe UNDER [bumed himself, Home then held it withi Month Sieone Tiiestiones. MeL \\
        \bottomrule
    \end{tabularx}
    \label{tab:burnt_hands}
\end{table}

\subsection{End-to-end models}

Having looked at the modular pipelines, we now compare the three end-to-end models. We find that M.OCR has a slightly lower SpACER but higher CDD than OlmOCR, whilst Granite-Docling, a much smaller model, performs poorly, struggling with the complex degraded output. Granite has a much higher CDD than the other two models due to a much higher fraction of erroneous symbols generated during transcription.

What is surprising is that, excluding the degenerate parsers, the end-to-end models are all beaten by the pipeline models in terms of overall quality. On inspection, this is due to parsing issues in the end-to-end models, not present in the pipeline approach. 

Interestingly, due to their design goal of whole-page OCR, Granite and M.OCR both performed much worse when given ground-truth bounding boxes than when given the entire page. OlmOCR, in contrast, had an average $d_\textrm{ocr} = 0.005$, which is 20x lower than the page level $d_\textrm{total}$, and the lowest $d_\textrm{ocr}$ recorded.

\begin{table}[h]
\centering
\caption{Median SpACER and CDD by end-to-end model (d\_total)}
\label{tab:end_to_end}
\begin{tabular}{lcc}
\toprule
 Model  & SpACER & CDD \\
\midrule
Granite & 0.864 & 0.207 \\
M.OCR & 0.097 & 0.057 \\
OlmOCR & 0.106 & 0.044 \\
\bottomrule
\end{tabular}
\end{table}

\subsection{Low information analysis}

We see how setting a simple cutoff on $d_\textrm{ocr}$ can predict whether parsing or OCR is the bottleneck. To do this, we use all combinations of the parsing and OCR pipelines and analyse the results at the page level, giving 980 observations. Ignoring the issues with the smaller PPDoc models, we find a cutoff $\frac{d_\textrm{ocr}}{d_\textrm{total}}\geq 0.5$ has an F1 of 0.83. Removing the Trespassing models, the F1 is 0.91 as the data becomes more coherent. However, if we take a middle ground and assume the failure mode of the smaller PPDoc models is always parsing, then the F1 is 0.74.

\begin{figure}
    \centering
    \includegraphics[width=\linewidth]{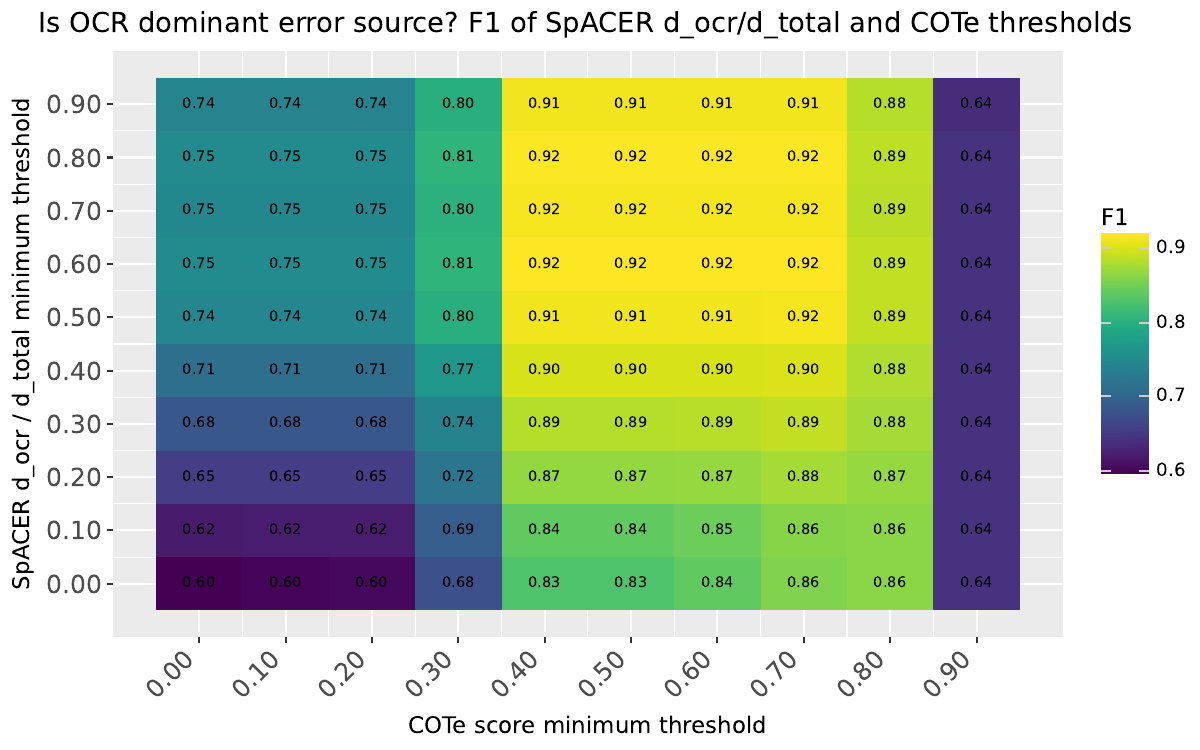}
    \caption{The figure shows that even conservative cutoffs for SpACER and COTe can identify if parsing or OCR quality is the main source of error in the pipeline}
    \label{fig:cutoff_thresholds}
\end{figure}

As the COTe score identified the Trespassing issue with the smaller PPDoc models, we include it in the analysis. We continue with the assumption that pipelines using the PPDoc models always have parsing as the dominant error source. We find that the interaction between the SpACER and COTe provides very high F1 for very conservative thresholds, for example, a threshold of $\frac{d_\textrm{ocr}}{d_\textrm{total}}\geq 0.5$ and $\textrm{COTe}\geq 0.5$, gives an F1 of 0.91. Details of the interaction can be seen in Figure \ref{fig:cutoff_thresholds}.

These findings suggest that using a cutoff of $\frac{d_\textrm{ocr}}{d_\textrm{total}}\geq 0.5$ is a reliable way to identify which error source is dominant. If the $d_\textrm{ocr}$ is combined with a COTe score threshold of 0.5, the result is even more informative and handles issues related to degenerate parsing gracefully, as the Trespass component detects predictions that erroneously span multiple ground truth regions.

\section{Discussion}

The CEV shows clear value as a global method for assessing the quality of character extraction. SpACER and the CDD correlate with CER when given ground-truth bounding boxes; they also correlate with the COTe score when analysing parse quality. SpACER also correlates with the CER when the approximate reading order is reconstructed, although the CDD, which measures shape change, has a lower correlation. 

Across all decompositions, we observe a CER-like tendency for SpACER to skew when aggregated using the mean, and we recommend the median as an alternative. We did not find evidence of hallucinating sentences; rather, in cases of large SpACER, sentences and phrases were mostly missed rather than inserted. This may be a result of using specialist OCR LLMs, as general-purpose models used for OCR have shown a tendency to hallucinate \cite{bourne_reading_2025}. Similar to \cite{bourne_cote_2026}, we find that the COTe score correctly interprets the model data relationship, whilst the F1/mAP fails to. This is a `pragmatic failure' of the F1, which is unable to handle the discrepancy in granularity between the ground truth and model predictions.

We also identify a pragmatic failure of the CEV, which occurs when the parsing models simply predict one or a small number of regions that cover the entire page. However, the initial anomalous readings in the total SpACER score can be traced to a parsing error in the decomposition, which was then identified as a major trespass across ground truth regions when combining SpACER with the COTe score. This means such errors can be filtered out automatically using thresholds that combine SpACER and COTe, mitigating the pragmatic failure through this combined approach. What is more, these SpACER/COTe thresholds can be used to identify whether the main source of error is due to parsing or OCR, even when character-level positions are not known. This reduces barriers to using the CEV and makes it compatible with the majority of document understanding datasets, which are comprised of groundtruth regions and associated text but not character or word positions.

The finding that even the large end-to-end models were outperformed by the smaller modular pipelines was certainly unexpected. It appears that the overall OCR quality of the models is very high, but their parsing ability is less capable. It is unclear whether this means that parsing and OCR are orthogonal, meaning that there is no loss in performance from a more transparent process, or simply that the end-to-end models have been trained on more limited datasets. The hypothesis that parsing and OCR are orthogonal is controversial, as a major motivator for end-to-end models is that the two components are mutually dependent. However, recent research has provided this hypothesis with support  \cite{cui_pp-ocrv5_2026}.

OCR approaches are increasingly being deployed within Digital Humanities as part of interpretive exercises, such as LLM text extraction and summarisation, which require interpretable benchmarks \cite{hemment_doing_2025}. We hope the Digital Humanities OCR specialist providers \cite{kahle_transkribus_2017, kiessling_escriptorium_2019, van_koert_loghi_2024} will include more interpretable and decomposable metrics as part of their platforms so as to encourage the active engagement of humanities researchers in data quality analysis.

Whilst this paper introduced the CEV as a general mechanism, it focused on the SpACER due to its ease of interpretation and similarity to CER. Further work could examine the CDD family of metrics to determine when approaches such as the JSD, Angular distance, Kullback-Leibler divergence, or others are most appropriate. Although we find useful heuristics to determine the main error source, another valuable area for further research is to develop a more mathematically principled approach to detecting the main source of error, such as exploiting the triangle inequality for true distance metrics.

\section{Conclusions}

The decomposable CEV metrics presented in this paper move beyond single-score benchmarking, enabling page parsing errors and character errors to be compared, separated, and disentangled in an intuitive manner, something not possible with the currently dominant CER approach. This is not to say that SpACER, the CER analogue, is a competitor; rather, in this paper, we have shown how an ecosystem of decomposable metrics, such as CER, the COTe score, and the CEV, complements each other. Such an ecosystem improves the interpretability of model performance, promoting the creation of robust and trustworthy models.

An irony of this paper is that \cite{bourne_cote_2026} gained additional insight into parsing model performance by shifting the focus from the physical position of text on the page to the narrative structure. In contrast, here we do the opposite, providing additional insight by shifting the focus of the CER's sequential narrative awareness to the physical positioning on the page.

In any case, SpACER, the CDD, and the CEV more broadly, provide a practical approach to evaluating OCR quality that is independent of any labelling schema. By integrating these measures into the COTe score framework and the \verb|cotescore| Python library, we provide practitioners with a valuable set of tools for building high-quality Document Understanding pipelines.

\section{Data Availability}

The code used for this project can be found at \url{https://github.com/JonnoB/SpACER}, The Python library \verb|cotescore| which includes the implementation of SpACER and the JSD version of the CDD can be found at \url{https://github.com/JonnoB/cotescore} or installed using `pip install cotescore'. The enriched version of the transcribed Spiritualist dataset can be found at \url{https://huggingface.co/datasets/Jonnob/the-spiritualist-enriched}

\printbibliography

\end{document}